\documentclass{article}

\usepackage{multirow}
\usepackage{graphicx}

\usepackage[preprint,nonatbib]{neurips_2022}

\usepackage[utf8]{inputenc} 
\usepackage[T1]{fontenc}    
\usepackage{url, booktabs, nicefrac, microtype, xspace}
\usepackage{graphicx, amsmath, amsfonts, amssymb, caption, subcaption, multirow, overpic, textpos, makecell}
\usepackage[table]{xcolor}
\usepackage[british, american]{babel}
\definecolor{citecolor}{HTML}{0071BC}
\definecolor{linkcolor}{HTML}{ED1C24}
\usepackage[pagebackref=false, breaklinks=true, letterpaper=true, colorlinks,
            citecolor=citecolor, linkcolor=linkcolor, bookmarks=false]{hyperref}
\usepackage[capitalize]{cleveref}

\newlength\savewidth\newcommand\shline{\noalign{\global\savewidth\arrayrulewidth
  \global\arrayrulewidth 1pt}\hline\noalign{\global\arrayrulewidth\savewidth}}

\renewcommand{\paragraph}[1]{\vspace{1.25mm}\noindent\textbf{#1}}

\newcolumntype{x}[1]{>{\centering\arraybackslash}p{#1pt}}
\newcolumntype{y}[1]{>{\raggedright\arraybackslash}p{#1pt}}
\newcolumntype{z}[1]{>{\raggedleft\arraybackslash}p{#1pt}}

\newcommand{\app}{\raise.17ex\hbox{$\scriptstyle\sim$}}

\definecolor{deemph}{gray}{0.6}

\definecolor{baselinecolor}{gray}{.9}
\newcommand{\baseline}[1]{\cellcolor{baselinecolor}{#1}}

\makeatletter
\DeclareRobustCommand\onedot{\futurelet\@let@token\@onedot}
\def\@onedot{\ifx\@let@token.\else.\null\fi\xspace}

\def\eg{\textit{e.g}\onedot} 
\def\ie{\textit{i.e}\onedot} 
 
 \def\vs{\textit{vs}\onedot}

\makeatother

\makeatletter\renewcommand\paragraph{\@startsection{paragraph}{4}{\z@}
	{.25em \@plus1ex \@minus.2ex}{-.5em}{\normalfont\normalsize\bfseries}}\makeatother

\title{Masked Autoencoders Enable Efficient  \\ Knowledge Distillers}

\author{
Yutong Bai\textsuperscript{1}\enspace
Zeyu Wang\textsuperscript{2}\enspace
Junfei Xiao\textsuperscript{1}\enspace
Chen Wei\textsuperscript{1} \\
\textbf{Huiyu Wang}\textsuperscript{1}\enspace 
\textbf{Alan Yuille}\textsuperscript{1}\enspace
\textbf{Yuyin Zhou}\textsuperscript{2}\enspace
\textbf{Cihang Xie}\textsuperscript{2}\\
\vspace{.5em}
\textsuperscript{1}Johns Hopkins University \enspace  \textsuperscript{2} University of California, Santa Cruz 
}

\begin{document}

\maketitle

\begin{abstract}

This paper studies the potential of distilling knowledge from pre-trained models, especially Masked Autoencoders. Our approach is simple: in addition to optimizing the pixel reconstruction loss on masked inputs, we minimize the distance between the intermediate feature map of the teacher model and that of the student model. This design leads to a computationally efficient knowledge distillation framework,  given 1) only a small visible subset of patches is used, and 2) the (cumbersome) teacher model only needs to be partially executed, \ie, forward propagate inputs through the first few layers, for obtaining intermediate feature maps. 

Compared to directly distilling fine-tuned models, distilling pre-trained models substantially improves downstream performance. For example, by distilling the knowledge from an MAE pre-trained ViT-L into a ViT-B, our method achieves 84.0\% ImageNet top-1 accuracy, outperforming the baseline of directly distilling a fine-tuned ViT-L by 1.2\%.
More intriguingly, our method can robustly distill knowledge from teacher models even with extremely high masking ratios: \eg, with 95\% masking ratio where merely TEN patches are visible during distillation, our ViT-B competitively attains a top-1 ImageNet accuracy of 83.6\%; surprisingly, it can still secure 82.4\% top-1 ImageNet accuracy by aggressively training with just FOUR visible patches (98\% masking ratio).
The code and models are publicly available at \url{https://github.com/UCSC-VLAA/DMAE}. 
\end{abstract}

\section{Introduction}
Following the success in the natural language processing \cite{Vaswani2017,devlin2018bert}, the Transformer architecture is showing tremendous potentials in computer vision \cite{vit,deit,cait}, especially when they are pre-trained with a huge amount of unlabelled data \cite{bao2021beit,zhou2021ibot}.
Masked image modeling, which trains models to predict the masked signals (either as raw pixels or as semantic tokens) of the input image, stands as one of the most powerful ways for feature pre-training. With the most recent representative work in this direction, masked autoencoder (MAE) \cite{he2021masked}, we are now able to efficiently and effectively pre-train high-capacity Vision Transformers (ViTs) with strong feature representations, leading to state-of-the-art solutions for a wide range of downstream visual tasks.

In this paper, we are interested in applying knowledge distillation \cite{hinton2015distilling}, which is one of the most popular model compression techniques, to transfer the knowledge from these strong but cumbersome ViTs into smaller ones.  
In contrast to prior knowledge distillation works \cite{hinton2015distilling,zhang2018deep,mirzadeh2020improved}, the teacher considered here is a pre-trained model whose predictions do not necessarily reveal the fine-grained relationship between categories; therefore, typical solutions like aligning the soft/hard logits between the teacher model and the student model may no longer remain effective. Moreover, after distilling the pre-trained teacher model, these student models need an extra round of fine-tuning to adapt to downstream tasks. These factors altogether turn distilling pre-trained models seemingly a less favorable design choice in terms of both performance and computational cost.

\begin{figure*}[t!]
    \centering
    \includegraphics[width=.98\linewidth]{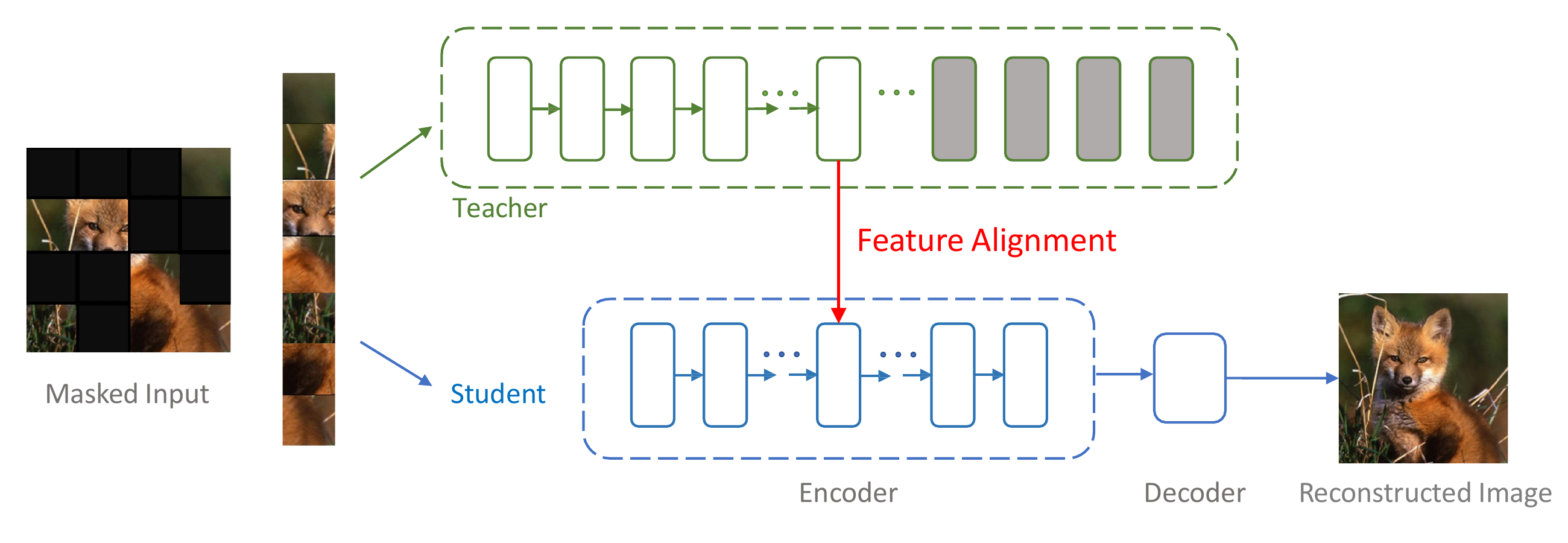}
    \vspace{-1.1em}
    \caption{\textbf{Illustration of the distillation process in DMAE.} 
    There are two key designs. Firstly, following MAE, we hereby only take visible patches as inputs and aims to reconstruct the masked ones. Secondly, knowledge distillation is achieved by aligning the intermediate features between the teacher model and the student model. Note the gray blocks denote the dropped high-level layers of the teacher model during distillation.} 
    \label{fig:method}
    \vspace{-1em}
\end{figure*}

Nonetheless, surprisingly, we find by building upon MAE, the whole distillation framework can efficiently yield high-performance student models. There are two key designs. Firstly, we follow MAE to let the encoder exclusively operate on a small visible subset of patches and to employ a lightweight decoder for pixel reconstruction. Whereas rather than using the ``luxury'' setups in MAE, we show aggressively \textit{simplifying pre-training from 1600 epochs to 100 epochs} and \textit{pushing masking ratio from 75\% to 95\%} suffice to distill strong student models. Secondly, instead of aligning logits, we alternatively seek to match the intermediate feature representation; this enables the cumbersome teacher model to only forward propagate inputs through the first few layers, therefore, reducing computations. We note applying \textit{a multi-layer perceptron (MLP) for feature projection} and \textit{L1 norm for distance measure} are essential recipes for ensuring a successful intermediate feature alignment.

We name this distilling MAE framework as DMAE. Compared to the traditional knowledge distillation framework where the teacher is a fine-tuned model, DMAE is more efficient and can train much stronger student models at different capacities. For example, by setting ViT-B as the student model, while the baseline of distilling a fine-tuned ViT-L achieves 82.8\% top-1 ImageNet accuracy, DMAE substantially boosts the performance to 84.0\% (+1.2\%) top-1 ImageNet accuracy, at a even lower training cost (\ie, 195 GPU hours \vs 208 GPU hours, see Table \ref{tab:cost}).
More intriguingly, we found that DMAE allows for robust training with extremely highly masked images---even with TEN visible patches (\ie, 95\% masking ratio), ViT-B can competitively attain a top-1 ImageNet accuracy of 83.6\%; this masking ratio can further be aggressively pushed to 98\% (FOUR visible patches) where DMAE still help ViT-B secure 82.4\% top-1 ImageNet accuracy. We hope this work can benefit future research on efficiently unleashing the power of pre-train models.

\section{Related Work}

\paragraph{Knowledge distillation (KD)} is a popular model compression technique that allows models to achieve both strong performances of large models and fast inference speed of small models. The first and seminal KD approach, proposed in~\cite{hinton2015distilling}, transfers the ``dark knowledge'' via minimizing the KL divergence between the soft logits of the teacher model and that of the student model.
From then on, many advanced KD methods have been developed, which can be categorized into two branches: logits distillation~\cite{furlanello2018born,zhang2018deep,cho2019efficacy,mirzadeh2020improved,zhao2022decoupled} and intermediate representation distillation~\cite{romero2014fitnets,kim2018paraphrasing,heo2019knowledge,heo2019comprehensive,Tian2020Contrastive}. Our proposed DMAE belongs to the second branch, as it minimizes the distance between latent features of the teacher model and those of the student model.

The first feature-based distillation method is FitNets~\cite{romero2014fitnets}. In addition to aligning logits, FitNets requires the student model to learn an intermediate representation that is predictive of the intermediate representations of the teacher network. 
Heo et al.~\cite{heo2019comprehensive} re-investigates the design of feature distillation and develops a novel KD method to create a synergy among various aspects, including teacher transform, student transform, distillation feature position, and distance function. CRD~\cite{Tian2020Contrastive} incorporates contrastive learning into KD to capture correlations and higher-order output dependencies. Unlike these existing works, our DMAE is the first to consider applying KD to extra information from self-supervised pre-trained models.

\paragraph{Masked image modeling (MIM)} helps models to acquire meaningful representations by reconstructing masked images. The pioneering works are built on denoising autoencoders~\cite{vincent2010stacked} and context encoders~\cite{pathak2016context}. Following the success of BERT in natural language \cite{devlin2018bert}, and also with the recent trend of adopting Transformer \cite{Vaswani2017} to computer vision~\cite{vit}, there have emerged a set of promising works on applying MIM for self-supervised visual pre-training. BEiT~\cite{bao2021beit} first successfully adopts MIM to ViT pre-training by learning to predict visual tokens. MaskFeat~\cite{wei2021masked} finds that learning to reconstruct HOG features enables effective visual representation learning. SimMIM~\cite{xie2021simmim} and MAE~\cite{he2021masked} both propose to directly reconstruct the pixel values of the masked image patches. Our work is built on MAE and finds that MAE enables the whole distillation framework to be efficient and effective.

\section{Approach}
\subsection{Masked Autoencoders}
Our method is built upon MAE, a powerful autoencoder-based MIM approach. Specifically,  the MAE encoder first projects unmasked patches to a latent space, which are then fed into the MAE decoder to help predict pixel values of masked patches. The core elements in MAE include:

\paragraph{Masking.} MAE operates on image tokens, \ie, the image needs to be divided into non-overlapping patches. A random small subset of those patches will be kept for the MAE encoder, and the rest will be set as the predicting target of the MAE decoder. Typically, a high masking ratio (\eg,75\%) is applied, preventing models from taking shortcuts (\eg, simply extrapolating missing pixels based on neighbors) in representation learning.

\paragraph{MAE encoder.}The MAE encoder is a standard ViT architecture except that it only operates on those unmasked patches. This design largely reduces the computation cost of encoders.

\paragraph{MAE decoder.} In addition to the encoded features of unmasked patches (from MAE encoder), the MAE decoder receives mask tokens as input, a learned vector shared across all missing positions. The mask token is only used during pre-training, allowing independent decoder design. Particularly, MAE adopts a lightweight decoder for saving computations.

\paragraph{Reconstruction.} Different from BEiT~\cite{bao2021beit} or MaskFeat \cite{wei2021masked}, MAE directly reconstructs image pixel values. The simple mean squared error is applied to masked tokens for calculating loss.

Note that, other than distillation-related operations, the whole pre-training and fine-tuning process in this paper exactly follow the default setup in MAE, unless specifically mentioned. Interestingly, compared to MAE, our DMAE robustly enables a much more efficient pre-training setup, \eg, 100 (\vs 1600) training epochs and 95\% (\vs 75\%) masking ratio.

\subsection{Knowledge Distillation}
MAE demonstrates extraordinary capabilities in learning high-capacity models efficiently and effectively. In this work, we seek to combine knowledge distillation with the MAE framework, to efficiently acquire small and fast models with similar performance as those powerful yet cumbersome models. The most straightforward approach is directly applying existing knowledge distillation methods, like the one proposed in DeiT \cite{deit}, to a fine-tuned MAE model. However, we empirically find that this approach hardly brings in improvements. In addition, this approach fails to leverage the special designs in MAE for reducing computations, \eg, only feeding a small portion of the input image to the encoder. To this end, we hereby study an alternative solution: directly applying knowledge distillation at the pre-training stage.

Since there are no categorical labels in MAE pre-training, distilling logits can hardly help learn semantically meaningful representations. We, therefore, resort to distilling the intermediate features. This idea is first developed in FitNets~\cite{romero2014fitnets}, and inspired a set of followups for advancing knowledge distillation \cite{kim2018paraphrasing,heo2019knowledge,heo2019comprehensive,Tian2020Contrastive}. Concretely, we first extract the features from the specific layers of the student model; after feeding such features into a small project head, the outputs will be asked to mimic the features from the corresponding layers of the teacher model. In practice, the projection head is implemented by a two-layers MLP, which not only addresses the possible feature dimension mismatch between teacher models and student models, but also allows extra flexibility for student models in learning from teacher models.

Formally, let $\mathbf{x} \in \mathbb{R}^{3 H W \times 1}$ be the input pixel RGB values and $\mathbf{y} \in \mathbb{R}^{3 H W \times 1}$ be the predicted pixel values, where $H$ denotes image height and $W$ denotes image width. The MAE reconstruction loss $L_{MAE}$ can be written as
\begin{equation}
L_{MAE}=\frac{1}{\Omega\left(\mathbf{x}_{M}\right)} \sum_{i \in M}\left(\mathbf{y}_{i}-\mathbf{x}_{i}\right)^{2}.
\label{equation: mae reconstruction loss}
\end{equation}
where $M$ denotes the set of masked pixels, $\Omega(.)$ is the number of elements, and $i$ is the pixel index.

Let $\mathbf{z}_{l}^{S},\mathbf{z}_{l}^{T} \in \mathbb{R}^{LC \times 1}$ be the features extracted from the $l$th layer of the student model and the teacher model, respectively, where $L$ denotes the patch numbers, and $C$ denotes the channel dimension. We use $\sigma()$ to denote the projection network function. Our feature alignment distillation loss $L_{Dist}$ can be written as
\begin{equation}
L_{Dist}=\sum_{l} \frac{1}{\Omega\left(\mathbf{z}_{l}^{T}\right)} \sum_{i}\left\| \sigma\left(\mathbf{z}_{l}^{S}\right)_{i}-\mathbf{z}_{l,i}^{T} \right\|_{1}.
\label{equation: distillation loss}
\end{equation}

The final loss used in pre-training is a weighted summation of MAE reconstruction loss $L_{MAE}$ and the feature alignment distillation loss $L_{Dist}$, controlled by the hyperparameter $\alpha$:
\begin{equation}
L=L_{MAE} + \alpha \times L_{Dist}.
\label{equation: pretrain loss}
\end{equation}

The framework of DMAE is summarized in Figure~\ref{fig:method}. Following MAE, DMAE also takes masked inputs and performs the pretext task of masked image modeling. Besides, the corresponding features are aligned between the teacher model and the student model. It is worthy of highlighting that DMAE is an efficient knowledge distiller: 1) it only operates on a tiny subset of visible patches \ie, {a high masking ratio is applied}; and 2) aligning intermediate layer features reduce the computation cost of (cumbersome) teacher model. In the next section, we extensively compare our method with three baselines: the original MAE without any distillation, DeiT-style distillation, and feature alignment distillation in the supervised setting.

\section{Experiments}
\label{sec:experiments}

\subsection{Implementation Details}
\label{sec:implementation details}
Following MAE~\cite{he2021masked}, we first perform self-supervised pre-training on ImageNet-1k~\cite{deng2009imagenet}.  Unless otherwise mentioned, the teacher models are public checkpoints released from the official MAE implementations\footnote{https://github.com/facebookresearch/mae}. For pre-training, we train all models using AdamW optimizer~\cite{loshchilov2017decoupled}, with a base learning rate of 1.5e-4, weight decay of 0.05, and optimizer momentum $\beta_1,\beta_2=0.9,0.95$. We use a total batch size of 4096, and pre-train models for 100 epochs with a warmup epoch of 20 and a cosine learning rate decay schedule. We by default use the masking ratio of 75\%; while the ablation study shows that our method can robustly tackle extremely high masking ratios. \emph{After pre-training, the teacher model is dropped, and we exactly follow the default setups in MAE to fine-tune the student model on ImageNet}.

For feature alignments, we choose to align the features from the $\frac{3}{4}$ depth of both the student model and the teacher model, which we find delivers decent results for all model sizes tested. For example, with a 24-layer ViT-L as the teacher model and a 12-layer ViT-B as the student model, features from the 9th layer of ViT-B are aligned with the features from the 18th layer of ViT-L. For feature projection, we set the hidden dimension of the two-layers MLP to be the same as the dimension of the aligned features from the teacher model. A GELU activation function~\cite{hendrycks2016gaussian} is appended in this two-layer MLP. We set $\alpha=1$ in Eq.~\ref{equation: pretrain loss} to balance the tradeoff between MAE reconstruction loss $L_{MAE}$ and the feature alignment distillation loss $L_{Dist}$ in pre-training.

\subsection{Analysis}
We first provide a detailed analysis of how to set distillation-related parameters in DMAE. Specifically, we set the teacher model as an MAE pre-trained ViT-L (from MAE official GitHub repository, attaining 85.9\% top-1 ImageNet accuracy after fine-tuning), and set the student model as a randomly initialized ViT-B. We analyze the following six factors:

\begin{table}[h]
\centering
\small
\begin{tabular}{l|c|c|c}
\shline
\# of Layers              & Layer Location          & Student Aligned Layer Index & ImageNet Top-1 Acc (\%)       \\ \shline
\multirow{4}{*}{Single}   & Bottom                  & 3                           & 82.6                      \\ \cline{2-4} 
                          & \multirow{2}{*}{Middle} & 6                           & 83.6                      \\ \cline{3-4} 
                          &                         & \baseline{9}                           & \textbf{\baseline{84.0}}             \\ \cline{2-4} 
                          & Top                     & 12                          & 83.4                      \\ \hline
Multiple & Middle+Top                 & 6+12                        & \textbf{84.2}                      \\ \hline
\end{tabular}
\vspace{.3em}
\caption{\textbf{The effects of feature alignment location.} We hereby test with 5 different layer locations, where top layers refer to those closer to network outputs. For single layer feature alignment, features from the $\frac{3}{4}$ depth of the model leads to the best ImageNet top-1 accuracy. This performance is even comparable to the setup of aligning multiple layers.}
\label{table: where to align}
\vspace{-1.5em}
\end{table}

\paragraph{Where to align.}
\label{paragraph: where to align}
In Table~\ref{table: where to align} we first check the effect of feature alignment location on model performance. We observe that shallower features are less favored: \eg, the 3rd layer alignment under-performs all other settings. We speculate this is due to the learning process of ViTs---images are much noisier and less semantic than texts, ViTs will first group the raw pixels in the bottom layers (closer to the input), which is harder to transfer. While features from $\frac{3}{4}$ depth (\ie, the 9th layer in ViT-B) achieves the best performance, a simple rule-of-thumb which we find fits all model scales in our experiments. We adopt this design choice in all other experiments. It is also worth mentioning that simply aligning multiple layers has no clear advantage over aligning features from $\frac{3}{4}$ depth (84.2\% \vs 84.0\%); we, therefore, stick to the $\frac{3}{4}$ depth setting, which is more efficient.

\paragraph{Aligning order.}
We next test the importance of alignment ordering on model performance. Specifically,
by fixing the layer location in the student model (\ie, the middle layer in our experiment), we then align it to different layers of the teacher model. As shown in Table \ref{tab:order}, we observe that when the aligned layers are in the same relative position (\eg, middle to middle), the student model can achieve the best performance.

\begin{table}[h!]
\centering
\small
\begin{tabular}{l|c}
\shline
Teacher Layer (relative position) & ImageNet Top-1 Acc (\%) \\ \shline
\baseline{Middle}                                          & \baseline{84.0}              \\ 
Top                                          & 83.3              \\ 
Bottom                                       & 82.1             \\ \hline
\end{tabular}
\vspace{.3em}
\caption{The analysis of aligning order.}
\label{tab:order}
\vspace{-1.5em}
\end{table}

\begin{figure}[h!]
    \centering
    \vspace{-1em}
    \includegraphics[width=\linewidth]{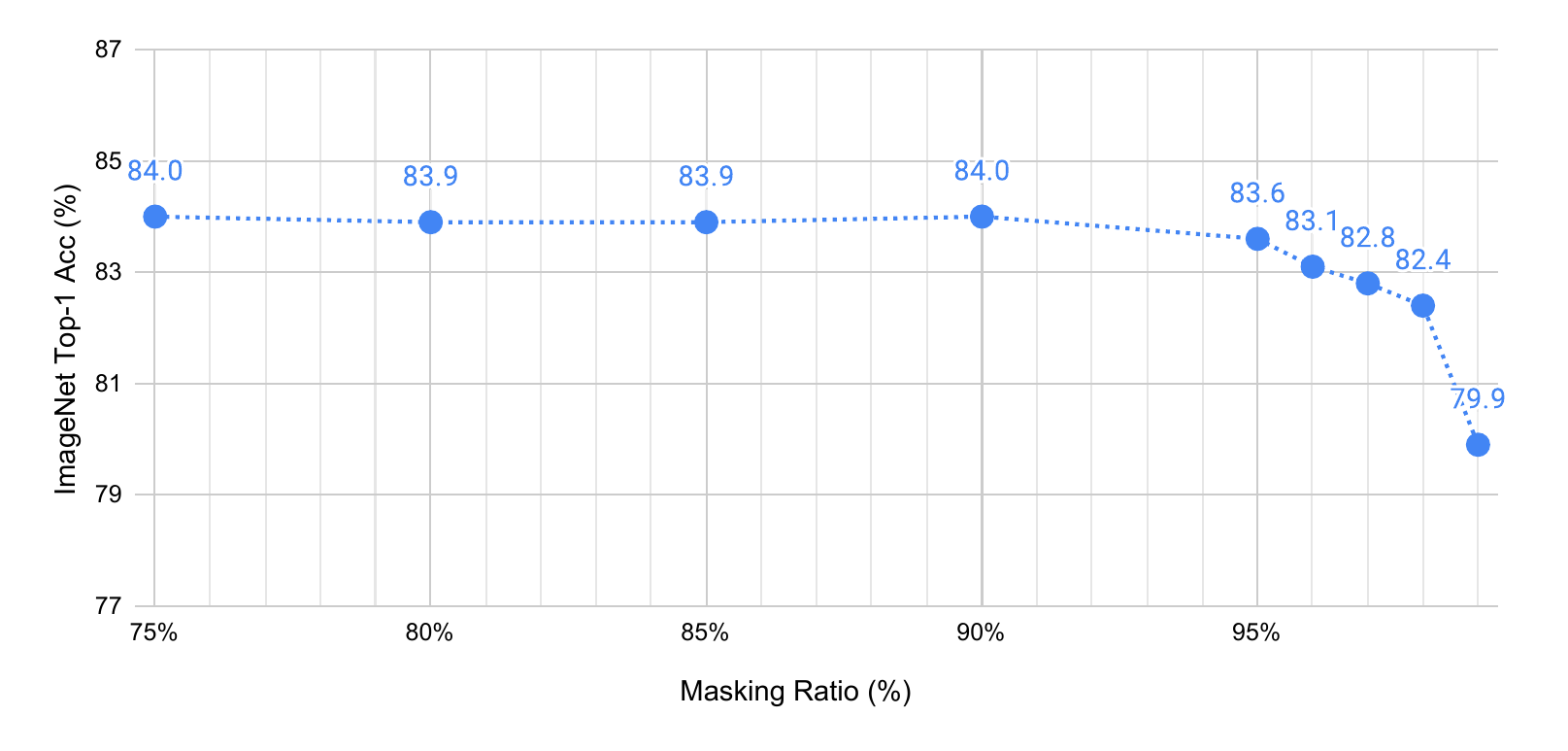}
    \vspace{-2em}
    \caption{\textbf{DMAE allows an extremely high masking ratio.} From left to right, we increase the masking ratio from the basic 75\% to the extreme 99\%. We note that our DMAE competitively attains 83.6\% top-1 ImageNet accuracy with 95\% masking ratio (TEN visible patches), and still secures 82.4\% top-1 ImageNet accuracy even by learning with FOUR visible patches (98\% masking ratio).}
    \label{fig: masking ratio}
\end{figure}

\paragraph{Masking ratio.}
MAE reveals that the masking ratio in masked image modeling could be surprisingly high (75\%). The hypothesis is that by learning to reason about the gestalt of the missing objects and scenes, which cannot be done by extending lines or textures because of the high masking ratio, the model is also learning useful representations. Interestingly, we find that when combining MAE and knowledge distillation, an even much higher masking ratio is possible, as shown in Figure~\ref{fig: masking ratio}.

Firstly, it is interesting to note that, compared to the typical 75\% masking ratio setting, further raising the masking ratio to 90\% comes at no performance drop, \ie, both attain 84.0\% top-1 ImageNet accuracy. Next, even with an extremely large masking ratio like 98\% (only FOUR visible patches), DMAE still beats the 100-epoch MAE baseline that uses a masking ratio of 75\% (second row in Table~\ref{tab:baseline}), by a non-trivial-margin (82.4\% \vs 81.6\%). 
These results suggest that, with the assistance of distilled knowledge from the teacher model, the student model can make better use of visible patches, even at a very limited amount, for representation learning.

\paragraph{Projection head.}
The goal of the proposed feature alignment distillation is to encourage the student model to learn features that are predictive of features from a stronger teacher model. To that end, a small projection head is employed on features from the student model, to 1) project them onto a space of the same dimension as the hidden dimension of the teacher model, and 2) provide extra flexibility for feature alignment. We ablate the choice of this projection network, as shown in Table \ref{tab: projection network}. We can observe that applying a simple 2-layer MLP performs the best among other choices.

\paragraph{Decoder depth.}
In Table~\ref{tab:decoder depth} we analyze the effect of decoder depth. Similar to MAE \cite{he2021masked}, the final performance gets (slightly) increased with a deeper decoder. We choose a decoder depth of 8 as the default setting as in \cite{he2021masked}. Note that a decoder depth of 2 is also a competitive choice---compared to an 8-depth decoder, it significantly reduces the computation cost while only marginally sacrificing the accuracy by 0.3\%.

\paragraph{Loss designs.}
Table \ref{tab: loss function} ablates the loss design. 
While SimMIM \cite{xie2021simmim} shows that L1 distance and L2 distance lead to similar performance, ours suggests that L1 distance exhibits a clear advantage over L2 distance, \ie, +0.7\% improvement. Furthermore, we note DMAE is quite robust to the specific value of the hyperparameter $\alpha$, which controls the relative importance of the distillation loss over the reconstruction loss. Based on these results, we choose L1 in Eq.~\ref{equation: distillation loss} for distance measure and set $\alpha=1$ in Eq.~\ref{equation: pretrain loss} for the rest experiments.

\begin{table}[h!]
\centering
\subfloat[
\textbf{Projection Head.} A simple 2-layer MLP works the best. We choose this as the default setting.
\label{tab: projection network}
]{%
\small
\begin{minipage}{0.45\linewidth} 
\begin{center}
\begin{tabular}{y{57}|x{97}}  \shline
Projection Head     & ImageNet Top-1 Acc (\%)  \\ \shline
Linear      & 83.6 \\
\baseline{2-layer MLP} & \baseline{84.0} \\
3-layer MLP & 83.8 \\ \hline
\end{tabular}
\end{center}
\end{minipage}
}
\hfill
\subfloat[
\textbf{Decoder Depth.} A deeper decoder (slightly) improves the pre-trained representation quality.
\label{tab:decoder depth}
]{%
\begin{minipage}{0.45\linewidth} 
\begin{center}
\small
\begin{tabular}{y{57}|x{97}} \shline
Decoder Depth & ImageNet Top-1 Acc (\%) \\ \shline
2 &  83.7 \\
4 &  83.8  \\
\baseline{8} & \baseline{84.0}  \\ \hline
\end{tabular}
\end{center}
\end{minipage}
}
\\
\subfloat[
\textbf{Loss Function.} From the first block, we can observe that L1 distance yields significantly higher performance than L2 distance. From the second block, we can observe that the hyperparameter $\alpha$ has little influence on the representation quality of DMAE.
\label{tab: loss function}
]{
\begin{minipage}{0.9\linewidth}

\begin{center}
\small
\begin{tabular}{y{80}|x{100}|x{100}} \shline
 & Loss Design& ImageNet Top-1 Acc (\%) \\ 
\shline
\multirow{2}{*}{Loss Choice} & \baseline{L1 with $\alpha=1$} &\baseline{84.0} \\
&L2 with $\alpha=1$ &83.3 \\
\hline
&L1 with $\alpha=0.5$ & 83.9 \\
\multirow{2}{*}{Loss Ratio} & \baseline{L1 with $\alpha=1$} & \baseline{84.0}  \\
&L1 with $\alpha=2$ & 84.0  \\
&L1 with $\alpha=4$ & 84.0  \\ \hline
\end{tabular}
\end{center}
\end{minipage}
}
\caption{The ablation studies on project head, decoder depth and loss function.}
\vspace{-1em}
\end{table}

\subsection{Comparison with Baselines.}
\label{sec:baselines}
In Table \ref{tab:baseline}, we compare the performance of our DMAE with various baselines: 

\paragraph{MAE.} The MAE baselines are presented in the first block of Table \ref{tab:baseline}. If MAE is also asked to pre-train for only 100 epochs, DMAE can substantially outperform this baseline by 2.4\% (from 81.6\% to 84.0\%). When comparing to a much stronger but more computationally expensive MAE baseline with 1600 pre-training epoch, we note DMAE still beats it by 0.4\%.

\paragraph{Supervised model.} The second block in Table \ref{tab:baseline} demonstrates the effectiveness of DMAE compared with models trained under the supervision of categorical labels, which requires a much longer training time, \ie, +2.2\% compared to the DeiT 300 epochs supervised training.

\paragraph{Other distillation strategies.} We next compare DMAE with other distillation methods. We consider DeiT-style logit-based distillation \cite{deit}, CRD \cite{Tian2020Contrastive}, SRRL \cite{yang2021knowledge}, Dear-KD \cite{chen2022dearkd}, and feature alignment distillation.  Note that for these baselines, one significant difference from DMAE is that the student model here directly distill knowledge from a supervisely fine-tuned teacher model. Moreover, to make these baselines more competitive, the teacher models will first be MAE pre-trained and then fine-tuned on ImageNet-1k.

The results are shown in the third block of Table \ref{tab:baseline}. Firstly, we can observe that the DeiT-style logit-based distillation, either soft or hard, even hurts the student models' performance. This phenomenon potentially suggests that such a distillation strategy may not fit teacher models of ViT architectures. For other baselines, we note that feature alignment distillation performs the best; but this is still worse than DMAE (82.8\% \vs 84.0\%), indicating the importance and effectiveness of distilling knowledge from a pre-trained teacher model.

\begin{table}[!h]
\centering
\small
\begin{tabular}{l|cc|c}
\shline
Method                                    & \begin{tabular}[c]{@{}c@{}}Pre-training \\ epochs\end{tabular} & \begin{tabular}[c]{@{}c@{}}Supervised training / \\ fine-tuning epochs\end{tabular} & \begin{tabular}[c]{@{}c@{}}ImageNet \\ Top-1 Acc (\%)\end{tabular}\\ \shline
MAE-B                                  & 100              & 100                               & 81.6         \\
MAE-B                                  & 1600             & 100                               & 83.6         \\ \hline
DeiT-B                                 & -               & 100                               & 76.8         \\
DeiT-B                                 & -                & 300                               & 81.8         \\ \hline
DeiT-B-Soft Distillation                    & -                & 100                               & 77.5         \\
DeiT-B-Hard Distillation                    & -                & 100                               & 78.3         \\ 
CRD\cite{Tian2020Contrastive}  & -                & 100                               &  81.9 \\
SRRL\cite{yang2021knowledge}  & -                & 100                               &  82.2 \\
Dear-KD\cite{chen2022dearkd}  & -                & 100                               &  82.4 \\
Supervised Feature Alignment & -                & 100                               & 82.8         \\ \hline
\baseline{DMAE-B}                                      & \baseline{100}              & \baseline{100}                               & \baseline{84.0}         \\ \hline
\end{tabular}
\vspace{.3em}
\caption{\textbf{DMAE shows stronger performance than all three kinds of baselines}: MAE, supervised model, and other existing advanced distillation strategies. }
\vspace{-1.5em}
\label{tab:baseline}
\end{table}

\subsection{Scaling to Different Model Sizes}
In this section, we test DMAE with different model sizes, listed in Table \ref{table: different model sizes}. For a fair comparison, both methods only pre-train models for 100 epochs. DMAE shows consistent improvement compared to MAE across different model sizes. With \textit{only one middle layer} feature alignment, DMAE brings an additional improvement of +2.4\% with ViT-B, +2.7\% with ViT-Small, and +3.4\% with ViT-Tiny. In addition, we are interested in the following two cases: 
    
\paragraph{Same teacher model, different student model.}
As shown in the first two lines in Table \ref{table: different model sizes}, we find that when using ViT-L as the teacher model, both ViT-B and ViT-S benefit from the distillation, demonstrating a clear advantage over the MAE baseline. We argue that the ability to effectively generalize to cases where an even smaller student model is desirable, especially for those computation-constrained real-world applications.
    
\paragraph{Same student model, different teacher model.} 
As shown in Table \ref{table: different model sizes}, with a ViT-S as the student model, enlarging the teacher model from ViT-B to ViT-L further boosts the accuracy by 0.8\% (from 79.3\% to 80.1\%). This result suggests that DMAE can effectively distill knowledge from the teacher models at different scales.

\begin{table}[h!]
\centering
\small
\begin{tabular}{c|c|cc}
\shline
\multicolumn{1}{l|}{\multirow{2}{*}{Student Model}} & \multicolumn{1}{l|}{\multirow{2}{*}{Teacher Model}} & \multicolumn{2}{c}{ImageNet Top-1 Acc (\%)}            \\ \cline{3-4} 
\multicolumn{1}{l|}{}                               & \multicolumn{1}{l|}{}                               & \multicolumn{1}{l|}{MAE} & DMAE        \\ \shline
Base                                                 & Large                                               & \multicolumn{1}{c|}{81.6}                 & 84.0 \textcolor{red}{(+2.4)} \\ \hline
\multirow{2}{*}{Small}                               & Large                                               & \multicolumn{1}{c|}{77.4}                 & 80.1 \textcolor{red}{(+2.7)} \\ \cline{2-4} 
                                                     & Base                                                & \multicolumn{1}{c|}{77.4}                 & 79.3 \textcolor{red}{(+1.9)} \\ \hline
Tiny                                                 & Base                                                & \multicolumn{1}{c|}{66.6}                 & 70.0 \textcolor{red}{(+3.4)} \\ \hline
\end{tabular}
\vspace{.3em}
\caption{DMAE shows consistent improvements compared with MAE across different model sizes.}
\vspace{-1.5em}
\label{table: different model sizes}
\end{table}

\subsection{Limited Training Data}
In certain real-world applications, data could be hard to acquire because of high data collection and labeling costs or due to privacy concerns. 
Leveraging models pre-trained on large-scale unlabeled datasets for fine-tuning when only a small dataset of downstream tasks is available becomes a promising solution. 
Here we seek to test the potential of our DMAE in this data-scarce scenario. We strictly follow \cite{chen2020simple} to sample 1\% or 10\% of the labeled ILSVRC-12 training datasets in a class-balanced way. We set MAE pre-trained ViT-L as the teacher model and a randomly initialized ViT-B as the student model. In addition, we compare DMAE with three kinds of baselines described in Section \ref{sec:baselines}: MAE, supervised model, and other distillation strategies, and similarly, set the teacher model to be a ViT-L that first pre-trained with MAE and then fine-tuned on ImageNet-1k.  Note that since DMAE has full access to the 100\% ImageNet dataset (without labels) during pre-training, to ensure a fair and competitive comparison,  \emph{we initialize all the baselines as the 1600-epoch MAE pre-trained model on ImageNet}.

Table~\ref{tab:baseline pct} shows that DMAE largely surpasses all other baselines. For example, when only 10\% ImageNet data is available for supervised training or fine-tuning, DMAE outperforms the MAE pre-trained baseline by 8.4\% (\ie, 73.4\% \vs 65.0\%). DMAE also significantly outperforms other distillation strategies, with an improvement ranging from 5.8\% to 7.0\%. We note this accuracy gap is even larger when only 1\% ImageNet is available, demonstrating the data efficiency of DMAE.

\begin{table}[!h]
\centering
\small
\begin{tabular}{l|cc|c|c}
\shline
Method                                                                              & \begin{tabular}[c]{@{}c@{}}Pre-training \\  epochs\end{tabular} & \begin{tabular}[c]{@{}c@{}}Supervised training / \\ fine-tuning epochs\end{tabular} & \begin{tabular}[c]{@{}c@{}}IN-1\% \\  Top-1 Acc(\%)\end{tabular}& \begin{tabular}[c]{@{}c@{}}IN-10\% \\  Top-1 Acc(\%)\end{tabular}\\ \shline
MAE-B                                                                            & 100              & 100                                                                            & 33.9             & 65.0              \\
MAE-B                                                                            & 1600             & 100                                                                            &  49.6                &  72.8                 \\ \hline
DeiT-B                                                                           & -                & 100                                                                            & -               &   -               \\ \hline
DeiT-B-Soft Distillation                                                              & 1600                & 100                                                                            & 36.0             & 66.4              \\
DeiT-B-Hard Distillation                                                              & 1600                & 100                                                                            & 37.3             & 67.3              \\
Supervised Feature Alignment & 1600              & 100                                                                            & 34.2             & 67.6              \\ \hline
\baseline{DMAE-B}                                                                                & \baseline{100}              & \baseline{100}                                                                            & \baseline{50.3}    &  \baseline{73.4}     \\ \hline
\end{tabular}
\vspace{.3em}
\caption{\textbf{DMAE demonstrates much stronger performance than all other baselines when training data is limited.} Note that the DeiT-B baseline is unable to converge because of data insufficiency. }
\vspace{-1.5em}
\label{tab:baseline pct}
\end{table}

\subsection{Computational Costs}
In Table~\ref{tab:cost}, we provide a quantitative evaluation on the computational cost, which is tested on a single NVIDIA A5000 GPU. We can observe that, without a  significantly increasing of training hours, DMAE substantially outperforms the MAE-100 baseline by +2.4\% on ImageNet; this result even exceeds the MAE-1600 baseline, which causes \app7x GPU hours than MAE-100 baseline. We additionally provide the actual GPU hours of other baselines, and find that the proposed DMAE stands as the most efficient one, meanwhile achieving the best top-1 ImageNet accuracy.

\begin{table}[h]
\centering
\small
\begin{tabular}{l|llcc}
\shline
\multicolumn{1}{l|}{\multirow{2}{*}{Model}} & \multicolumn{3}{c|}{Training Cost (GPU Hours)}                                            & \multicolumn{1}{c}{ImageNet}                                                   \\ \cline{2-4}
\multicolumn{1}{c|}{}                       & Pre-training & \multicolumn{1}{|l|}{Fine-tuning} & \multicolumn{1}{c|}{Overall} & Top-1 Acc(\%) \\ 
\shline
MAE-B-100-epoch                            &\multicolumn{1}{c|}{78h}               & \multicolumn{1}{c|}{112h}           & \multicolumn{1}{c|}{190h}      & 81.6     \\
MAE-B-1600-epoch                           &\multicolumn{1}{c|}{1248h}             & \multicolumn{1}{c|}{112h}           & \multicolumn{1}{c|}{1360h}     & 83.6     \\
DeiT-B-Soft Distillation                     &\multicolumn{1}{c|}{-}             & \multicolumn{1}{c|}{213h}           & \multicolumn{1}{c|}{213h} & 77.5     \\
DeiT-B-Hard Distillation                     &\multicolumn{1}{c|}{-}             & \multicolumn{1}{c|}{213h}           & \multicolumn{1}{c|}{213h}    & 78.3     \\ 
Supervised Feature Alignment                       &\multicolumn{1}{c|}{-}             & \multicolumn{1}{c|}{208h}           & \multicolumn{1}{c|}{208h}    & 82.8     \\ \hline
DMAE-B                                  &\multicolumn{1}{c|}{83h}              & \multicolumn{1}{c|}{112h}           & \multicolumn{1}{c|}{195h}   &  \textbf{84.0}  \\\hline  
\end{tabular}
\vspace{.3em}
\caption{\textbf{Computational cost comparisons among DMAE and other baselines.} The training cost is measured by A5000 GPU hours. We note the proposed DMAE maintains a similar (or even cheaper) training cost than others, while achieving much higher top-1 ImageNet accuracy.}
\label{tab:cost}
\vspace{-1.5em}
\end{table}

\subsection{Standard Deviation Analysis}
In the above experiments, we kept the same random seed. Following MAE, we perform the statistical analysis for DMAE by changing the random seeds. In Table \ref{tab:std}, from top to down, we show three aligning settings; and from left to right, we show the results with the default seed, the average accuracy with three randomly sampled seeds, and their standard deviation, respectively. From these results, we could conclude that our DMAE can bring in statistically stable improvements.

\begin{table}[h]
\centering
\small
\begin{tabular}{l|c|c|c}
\shline
Distillation Position (Student) & ImageNet Top-1 Acc(\%) & Avg over 3 times & Standard Deviation \\ \shline
Bottom (3th)                       & 82.6                                 & 82.50                     & 0.20                         \\ 
Mid (6th)                          & 83.6                                 & 83.67                     & 0.08                        \\ 
Top (9th)                          & 84.0                                 & 84.03                     & 0.10 \\ \hline                      
\end{tabular}
\vspace{.3em}
\caption{\textbf{Standard deviation analysis for DMAE.} From top to bottom, we show three aligning settings, the model performance with the default seed, the average performance with three randomly sampled seeds, and their standard deviations, respectively. \label{tab:std}}
\vspace{-1.5em}
\end{table}

\subsection{Generalization to Other Methods}
Lastly, we provide preliminary results of integrating DMAE into other self-supervised training frameworks, including DINO \cite{dino} and MoCo-V3 \cite{chen2021empirical}. Unlike MAE, which belongs to masked image modeling, DINO and MoCo-V3 are contrastive learning-based methods. Still, without further hyperparameter tuning, DMAE effectively shows non-trivial improvements on top of both DINO (+1.3\%) and MoCo-v3 (+1.4\%), demonstrating the potential of distilling pre-trained models (rather than fine-tuned models as in most existing knowledge distillation frameworks).

\begin{table}[!h]
\centering
\small
\begin{tabular}{l|c|c}
\shline
 & w/o DMAE & w/ DMAE        \\ \shline
DINO      & 80.9        & 82.2 \textcolor{black}{(+1.3)} \\ 
MoCo-V3   & 81.1        & 82.5 \textcolor{black}{(+1.4)} \\ \hline
\end{tabular}
\vspace{.3em}
\caption{DMAE effectively improves other self-supervised pre-training frameworks (including DINO and MoCo-v3) on ImageNet classification.}
\label{tab:general}
\vspace{-1em}
\end{table}

\section{Conclusion}
\label{sec:conclusion}
Self-supervised pre-training has demonstrated great success for those exponentially growing models in the natural language domain. Recently, the rise of MAE shows that a similar paradigm also works for the computer vision domain, and now the development of vision models may embark on a similar trajectory as in the language domain. Yet, it is often desirable to have a well-balanced model between performance and speed in real-world applications. This work is a small step towards unleashing the potential of knowledge distillation, a popular model compression technique, within the MAE framework. Our DMAE is a simple, efficient, and effective knowledge distillation method: feature alignment during MAE pre-training. Extensive experiments on multiple model scales demonstrate the effectiveness of our approach. Moreover, an intriguing finding is that it allows for a masking ratio even higher than the already large one used in MAE (\ie, 75\%). We have also validated the effectiveness of our DMAE when in the small-data regime. We hope this work can benefit future research in knowledge distillation with pre-trained models.

\section*{Acknowledgement}
This work is partially supported TPU Research Cloud (TRC) program, and Google Cloud Research Credits program.

{\small
\bibliographystyle{plain}
\bibliography{egbib}
}

\end{document}